# A Multi-Pass Large Language Model Framework for Precise and Efficient Radiology Report Error Detection

Songsoo Kim[1,2], Seungtae Lee[2], See Young Lee[3], Joonho Kim[1,4], Keechan Kan[5], Dukyong Yoon[1,6,7*]

[1]Department of Biomedical Systems Informatics, Yonsei University College of Medicine, Seoul, Republic of Korea

[2]Department of Radiology, Yonsei University College of Medicine, Seoul, Republic of Korea

[3]Department of Internal Medicine, Gangnam Severance Hospital, Yonsei University College of Medicine, Seoul, Republic of Korea

[4]Department of Neurology, Yonsei University College of Medicine, Seoul, Republic of Korea.

[5]Department of Surgery, Samsung Medical Center, Seoul, Republic of Korea

[6]Institute for Innovation in Digital Healthcare, Severance Hospital, Seoul, Republic of Korea

[7]Center for Digital Health, Yongin Severance Hospital, Yonsei University Health System, Yongin, Republic of Korea

*Corresponding author:

Dukyong Yoon: dukyong.yoon@yonsei.ac.kr; Department of Biomedical Systems Informatics, Yonsei University College of Medicine, 50-1 Yonsei-ro, Seodaemun-gu, Seoul, 03722, Republic of Korea, Phone number: 82-2-2228-2489

## Abstract

**Background**: The positive predictive value (PPV) of large language model (LLM) proofreaders is limited by the low prevalence of errors in radiology reports.

**Purpose**: To determine whether a three-pass LLM framework increases PPV and reduces operational costs compared with baseline approaches.

**Materials and Methods**: In this retrospective study, 1,000 consecutive radiology reports (radiography, ultrasonography, CT, and MRI; 250 each) were sampled from the Medical Information Mart for Intensive Care III database. Two public chest radiography corpora (CheXpert and Open-i) served as external test sets. Three LLM frameworks were evaluated: single-prompt detector; report extractor plus single-prompt detector; and extractor, detector, and false positive verifier. Precision for each framework was assessed using PPV and absolute true positive rate (aTPR). Overall efficiency was estimated as model inference charges plus reviewer remuneration. Pairwise differences were assessed using cluster bootstrap and exact McNemar tests; the Holm–Bonferroni method was used to control family-wise error rate.

**Results**: For precision, the PPV increased stepwise from 0.063 [95% CI, 0.036–0.101] in Framework 1 to 0.079 (0.049–0.118) in Framework 2 and 0.159 (0.090–0.252) in Framework 3 ($P < .001$ for Framework 3 vs both baselines). The aTPR remained stable (0.012–0.014, $P \geq .84$). Regarding efficiency, the cost per 1,000 reports decreased to USD 5.58 in Framework 3, compared with USD 9.72 in Framework 1 and USD 6.85 in Framework 2—representing cost reductions of 42.6% and 18.5%, respectively. The number of reports requiring human review decreased from 192 to 88. External validation showed the highest PPV for Framework 3 (0.133 in CheXpert and 0.105 in Open-i), with an unchanged aTPR of 0.007.

**Conclusion**: A three-pass LLM framework more than doubled precision and halved the cost of radiology report error detection without compromising error detection capability, offering sustainable strategies for AI-assisted quality assurance in routine practice.

**Introduction**

Large language models (LLMs) are being hailed as an additional set of eyes for proofreading radiology reports(1,2). However, when applied to real-world data, this extra eye often results in frequent false alarms. The precision of these models—also referred to as positive predictive value (PPV)—remains low because, despite "good" model specificity, the underlying error rate in clinical practice is extremely low. For example, in a setting with a 1% error prevalence, even a highly sensitive model with 90% specificity would still generate approximately 10 false alarms for every true error detected. In one experiment involving 10,000 real reports, GPT-4 achieved a PPV of only 6%, producing roughly 15 false alerts for each true error(3). These excessive notifications contribute to alert fatigue among radiologists, prompting them to ignore subsequent warnings, hindering effective human–AI collaboration, and—ironically—increasing the real-world workload(4). Although continued advances in LLMs are expected to address these shortcomings, the anticipated gains present a double-edged sword in terms of overall utility(5). Parameter scaling, task-specific fine-tuning, and deployment of multi-agent systems (6,7) can certainly enhance model performance and clinical efficiency. However, the trade-offs between improved performance and increased computational demands—along with their ecological footprint—remain largely unexamined. For instance, coordinating multiple agents routinely produces execution traces averaging more than 15,000 lines per session, while training a single GPT-3–scale model is estimated to emit over $6 \times 10^5$ kg $CO_2$-equivalents(8,9). Consequently, AI-driven radiology report error detection faces a dual imperative: it must increase precision to reduce human workload, while also maintaining computational and environmental efficiency.

Despite these limitations, previous studies still benchmark LLMs on error-inflated datasets and rarely explore strategies for improving PPV in low-error, real-world settings(1,10,11). Similarly, strategies to improve computational efficiency remain largely unexplored.

To address these gaps, we present a multi-pass LLM framework that improves both precision and efficiency. The pipeline (i) employs a lightweight report extractor to streamline LLM input, (ii) applies stepwise reasoning to improve error detection precision, and (iii) provides a user interface to facilitate rapid review of the model's structured output by radiologists. A benchmark with two non-optimized baselines was performed to quantify improvements in precision and efficiency.

**Materials and Methods**

This retrospective study was approved by the local institutional review board. Written informed consent was obtained from all participating human readers.

*Dataset Curation*

Radiology reports were retrieved from the Medical Information Mart for Intensive Care III (MIMIC-III) database(12). Modality-level stratified random sampling was performed to construct a balanced primary test set comprising 1,000 reports, with 250 reports each from radiography, ultrasonography, computed tomography, and magnetic resonance imaging. An additional hold-out set of 50 predominantly radiography reports was reserved exclusively for prompt tuning. To assess the external generalizability of the proposed pipeline, two publicly available radiology report datasets—CheXpert and Open-i Chest X-ray(13,14)—were used as external test sets (Figure 1).

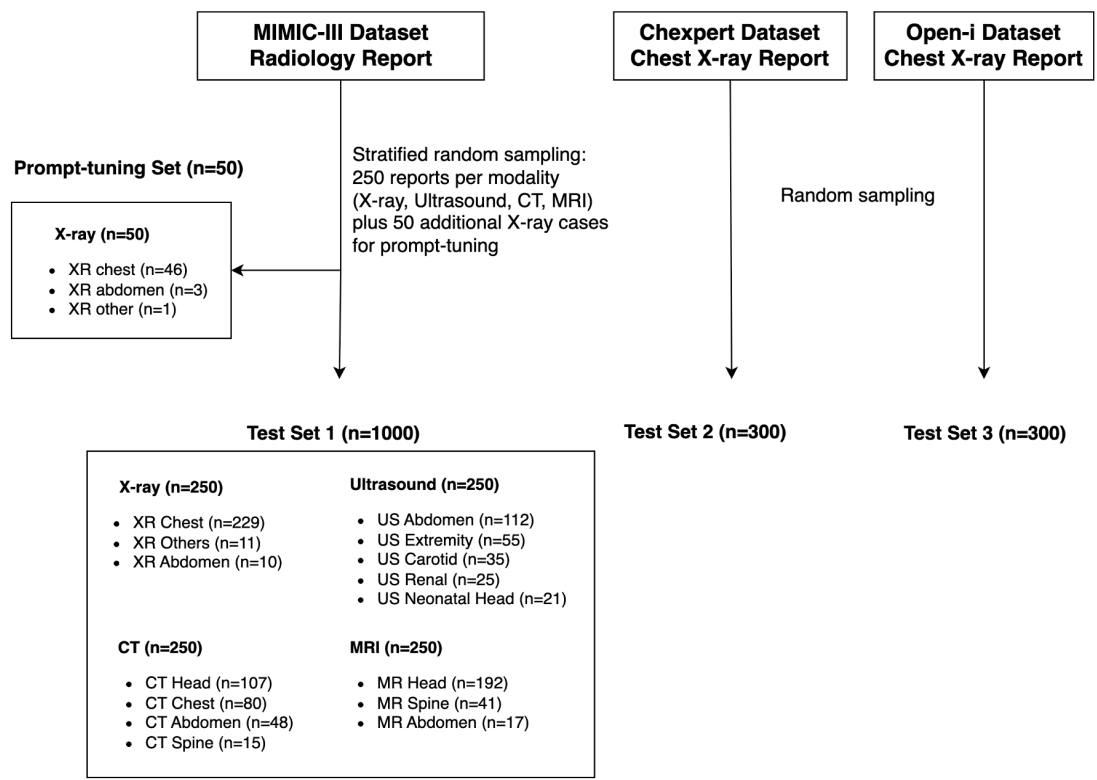

**Figure 1**. Flowchart of radiology report sampling from the MIMIC-III, CheXpert, and Open-i datasets for prompt tuning and test set construction

*Proposed Framework and Experimental Design*

Three LLM pipelines were compared (Figure 2). In Framework 1, the original report was input directly into an advanced LLM, which performed both error detection and false positive (FP) verification within a single prompt. In Framework 2, a lightweight LLM first extracted and structured the relevant portion of the radiology report by removing content outside the Findings and Impression sections—such as clinical information, technique notes, and headers—and seamlessly merging any addenda into this section. The resulting structured Findings/Impression block was then passed to an advanced LLM, which performed combined error detection and FP verification in a single prompt. Framework 3 retained the preliminary extraction step but divided the downstream reasoning across two successive prompts: candidate errors were first enumerated and then re-examined to verify potential FPs.

All prompts were constructed with a task–context–output format structure. The required output format was enforced JSON Schema mode(15), which ensured that every response conformed to a predefined schema containing mandatory key pairs, such as findings/impression and error/error_reason. The resulting outputs were streamed to a web-based quality assurance interface, which displayed the flagged report alongside the model's error reasoning, allowing human reviewers to accept or reject each suggestion with a single click (Figure 3).

The lightweight LLM used in this experiment was executed by OpenAI's gpt-4.1-nano, selected for its favorable cost-effectiveness. The advanced LLMs were o3 and o4-mini, chosen for their superior reasoning performance at the time of study(16). All pipelines were executed on the institution's private Azure OpenAI Service, with each LLM API call launched within an isolated API session. Detailed descriptions of the prompts and parameters are provided in the supplementary material. All source code and the user interface, which enables modification of models and prompts, are openly available on GitHub (https://github.com/radssk/mp-rred).

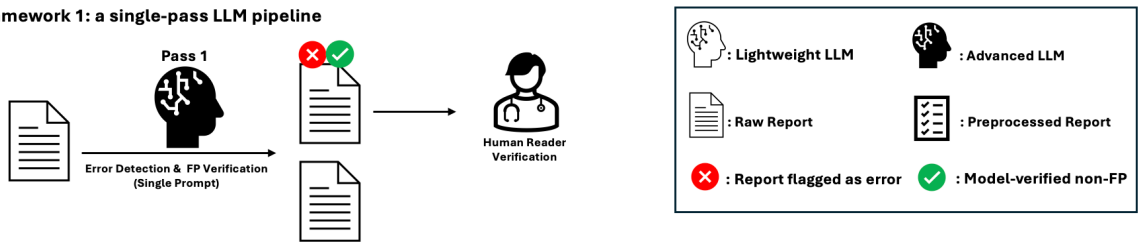

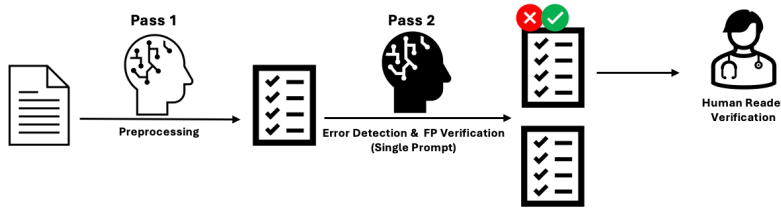

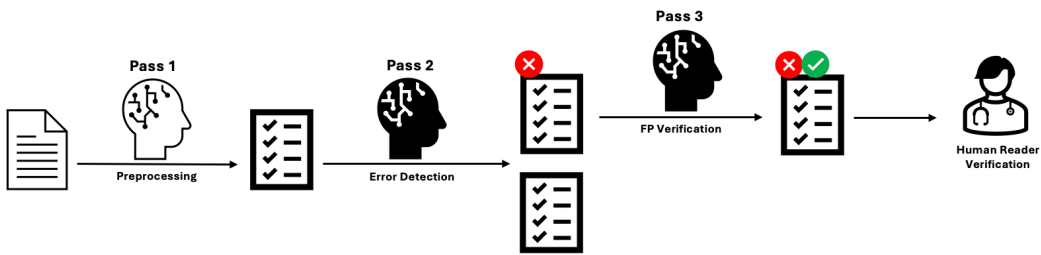

**Figure 2.** Experimental design of large language model pipelines for radiology report error detection. In the single-pass framework (a), each report is processed once by a lightweight LLM that simultaneously performs error detection and false positive (FP) verification before reader review. In the two-pass framework (b), a lightweight LLM first performs preprocessing, and an advanced LLM subsequently conducts combined detection and verification before reader review. In the proposed three-pass framework (c), preprocessing is followed by error detection in a second pass and isolated FP verification in a third pass by an advanced LLM, prior to reader review. Abbreviations: LLM = large language model, FP = false positive.

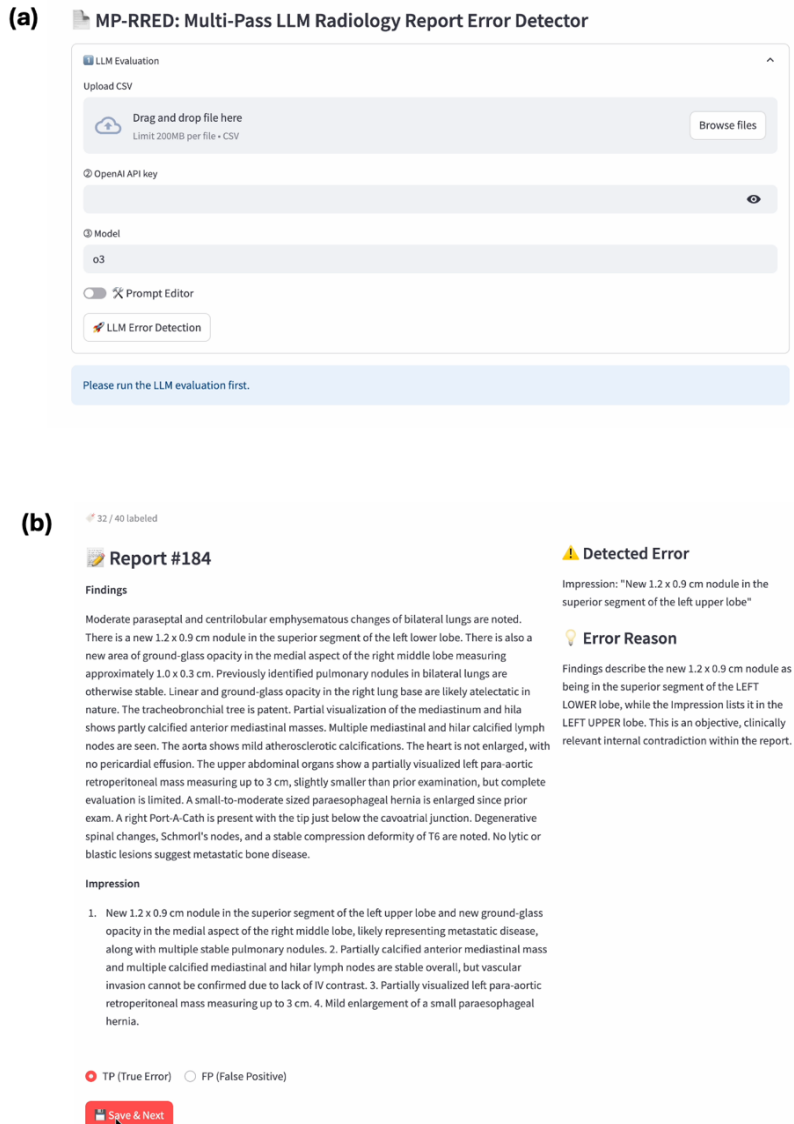

**Figure 3.** User interface for the multi-pass large language model radiology report error detector. (a) The evaluation console allows users to upload a CSV file of radiology reports, enter an OpenAI model and API key, and initiate error analysis. (b) The review screen loads the preprocessed JSON, displaying the Findings and Impression sections in the left panel, while the right panel shows the detected error and the model-provided rationale using the error and error_reason JSON keys. This structured layout enables reviewers to classify each finding as either a TP or a false positive (FP) with a single click. Abbreviations: LLM = large language model; TP = true positive; FP = false positive.

*Precision Evaluation*

Each flagged report was independently reviewed in two stages—first by one of two board-certified physicians and subsequently by one of two subspecialty-trained radiologists. The workload was equally distributed, and any disagreement was resolved by consensus. Performance of the framework was evaluated using PPV (PPV = TP / [TP + FP]) and the absolute true positive rate (aTPR = TP / N), where N denotes the size of the test set.

Here, true positive (TP) refers to a model-flagged report in which a genuine error was confirmed, while FP refers to a flagged report that did not contain a true error.

*Efficiency Evaluation*

Assuming a priori equivalence in absolute true positive rates across the three frameworks, a cost-minimization analysis was conducted, for which the primary outcome—total expenditure—was defined as the sum of (i) model-computational costs and (ii) reviewer labor costs (17).

Because the exact computational cost and carbon footprint of the closed-source OpenAI models could not be measured directly, we used per-token API charges as a surrogate indicator. This choice is supported by evidence that (i) electricity and GPU rental costs dominate token pricing in commercial LLMs(18,19) and (ii) inference energy consumption—and consequently $CO_2$ emissions—increases almost linearly with token count(20). Consequently, the LLM inference cost was obtained by multiplying the logged token count by the provider's per-token tariff (Supplementary Eq. S1). Reviewer labor cost was approximated by multiplying the total number of reports sent for manual inspection—comprising both true and false positives—by the mean remuneration paid per report (Supplementary Eq. S2). The total expenditure for each framework was therefore defined as the sum of the two components (Supplementary Eq. S3) and is reported separately to permit direct cost-effectiveness comparisons. Formal derivations and the full set of symbols are provided in Supplementary Materials.

*Statistical Analysis*

Continuous variables are reported as mean ± standard deviation when normally distributed (Shapiro–Wilk test, $p > 0.05$) and as median with interquartile range (IQR) otherwise. Categorical variables are summarized as counts and percentages. PPV and aTPR are expressed with two-sided 95% exact (Clopper–Pearson) confidence intervals(21).

For PPV comparisons, pairwise differences among the three frameworks were assessed using report-level paired-cluster bootstrap (10,000 replicates). Two-sided p-values were extracted from the bootstrap distributions, and the family-wise error rate across the three comparisons was controlled using the Holm–Bonferroni procedure(22). Modality-specific PPV analyses were regarded as exploratory and reported without multiplicity adjustment. When the frameworks followed a prespecified ordinal sequence, a Cochran–Armitage trend test was additionally applied to detect monotonic trends in PPV across the ordered groups.

For aTPR comparisons, within-case differences among the three frameworks were evaluated using the exact McNemar test, with family-wise error rate controlled via the Holm–Bonferroni procedure. When comparing three or more frameworks, an overall Cochran Q test was conducted; if significant, pairwise McNemar tests with Holm correction were performed. All tests were two-tailed, with α = 0.05.

The sample size was calculated based on the MIMIC-III dataset. The baseline PPV of the reference pipeline was assumed to be 6%, as reported previously(3). A two-fold improvement with the proposed pipeline was designated as the minimal clinically important difference. Treating the comparison as a two-sided test of the difference between two independent proportions and adopting α = 0.05 with a statistical power of 80%, a minimum of 716 reports was required. Consequently, the final sample of 1,000 reports satisfied and exceeded this requirement, thereby ensuring adequate power for the primary hypothesis test. Analyses were performed in Python 3.11 using pandas 2.2.2, SciPy 1.12.0 and statsmodels 0.15.0 for statistical procedures, and matplotlib 3.9.0 for visualization

## Results

*Detected Errors and False Positive Cases*

The true errors identified in each dataset are summarized in Supplementary Table 3. Fourteen errors were detected in the MIMIC-III dataset—two in chest radiographs, three in carotid ultrasonography studies, one in a neonatal brain ultrasonography study, three in head CT scans, two in chest CT scans, and three in head MRI examinations. Two errors were found in both the CheXpert and Open-i datasets.

Table 1 summarizes representative false-positive cases. FP that occurred only in Framework 1 arose chiefly from a rigid comparison of superficial header elements—such as date strings or minor omissions in the clinical history—with the body text, causing spurious contradiction flags. Once the header metadata had been removed, such false flags were not observed in Framework 2 or 3. Framework 2 still produced many FPs because it compared sentences at the strict word-level, with little regard for anatomical or contextual nuance. Framework 3 subsequently re-examined the candidate contradictions produced by Framework 2 against the full report; statements regarded as clinically acceptable in routine radiology were re-classified, and the overall FP burden was thereby reduced.

*Precision of LLM Frameworks*

The precision of the LLM frameworks improved as the pipeline complexity increased (Table 2, Figure 4A). In Framework 3, the overall PPV was 0.159 (95% CI: 0.090–0.252), compared with 0.079 in Framework 2 and 0.063 in Framework 1. The superiority of Framework 3 over both Framework 1 and Framework 2 remained significant after multiple comparison correction (all paired-cluster bootstrap $p < .001$; all Holm-adjusted $p < .001$). A prespecified Cochran–Armitage trend test confirmed a significant upward trend in PPV across the three ordered frameworks ($p = .019$), indicating that successive refinements effectively reduced FP alerts.

The observed increase in precision was not accompanied by a reduction in true positive detections (Table 3, Figure 4B). The overall aTPR was 0.014 (95% CI: 0.008–0.023) for Framework 3, compared with 0.013 for Framework 2 and 0.012 for Framework 1. None of the pairwise comparisons reached statistical significance (all $p \geq .84$), indicating that Framework 3 reduced FP flags without compromising error detection.

In the CheXpert and Open-i datasets, Framework 3 achieved the highest PPVs (0.133 and 0.105, respectively; paired-cluster bootstrap $p \geq .26$; Holm-adjusted $p \geq .79$) and maintained identical aTPRs across frameworks (0.007 for both datasets; all $p = 1.000$), demonstrating robustness across diverse datasets. However, in the Open-i dataset, Framework 2 yielded a lower PPV than Framework 1—the only instance in which PPV did not increase monotonically with pipeline complexity. This exception may be due to the Open-i dataset already being extensively preprocessed, reducing the relative benefit of the first-pass LLM.

When Framework 3 was executed using the o4-mini model instead of o3, the overall PPV significantly declined to 0.081 (p < .001; Supplementary Table 4), while the aTPR decreased slightly to 0.012 without reaching statistical significance (p = .693; Supplementary Table 5).

*Efficiency of LLM Frameworks*

The token counts for each pass are summarized in Table 4. Framework 3 incurred the lowest cost, at USD 5.58 per 1,000 reports, compared with USD 9.72 and USD 6.85 for Frameworks 1 and 2, respectively—corresponding to cost reductions of approximately 42.6% and 18.5% relative to Frameworks 1 and 2 (Figure 5). Framework 2 achieved most of its savings through token reduction via preprocessing, relative to Framework 1. In Framework 3, additional savings beyond those of Framework 2 were primarily attributed to the FP verifier being triggered for only 88 candidate errors, rather than for all cases. Assuming a human review fee of USD 3 per report, the total expenditure decreased from USD 585.66 in Framework 1 and USD 498.79 in Framework 2 to USD 269.52 in Framework 3—representing reductions of 54.0% and 45.8%, respectively.

A similar trend was observed in the CheXpert and Open-i datasets, where Framework 3 consistently demonstrated the lowest model inference costs per 300 reports (USD 0.1374 and USD 0.1271, respectively), compared with Framework 1 (USD 1.8943 and USD 1.5930, respectively). However, in the Open-i dataset, Framework 2 incurred a higher model inference cost than Framework 1—representing the only instance in which model inference cost did not decrease with increased pipeline complexity.

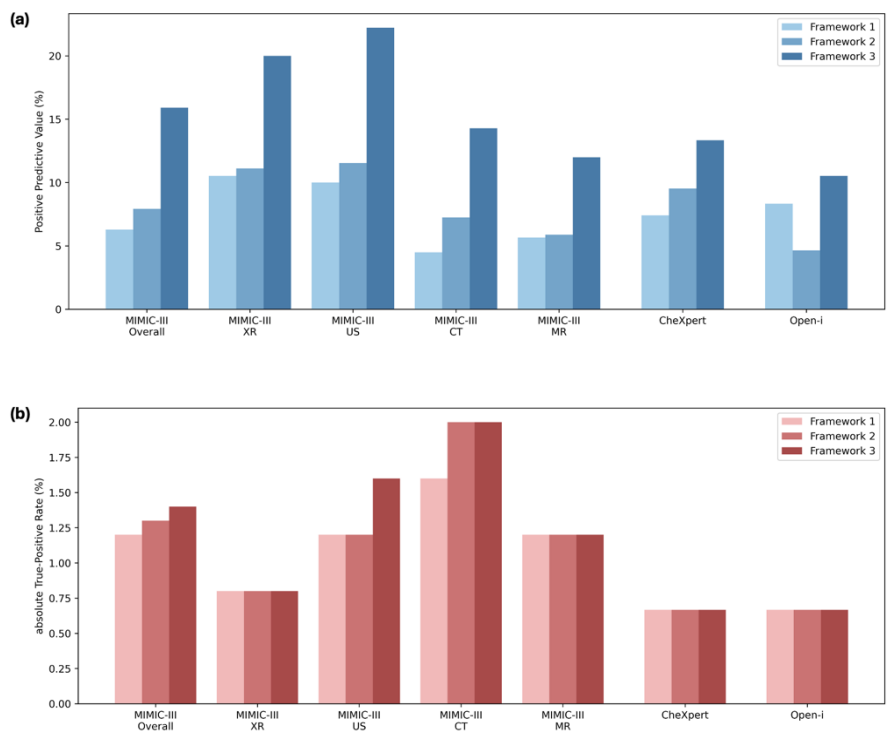

**Figure 4.** Performance comparison of the three error detection frameworks across the MIMIC-III, CheXpert, and Open-i datasets. (A) Positive predictive value. (B) Absolute true positive rate.

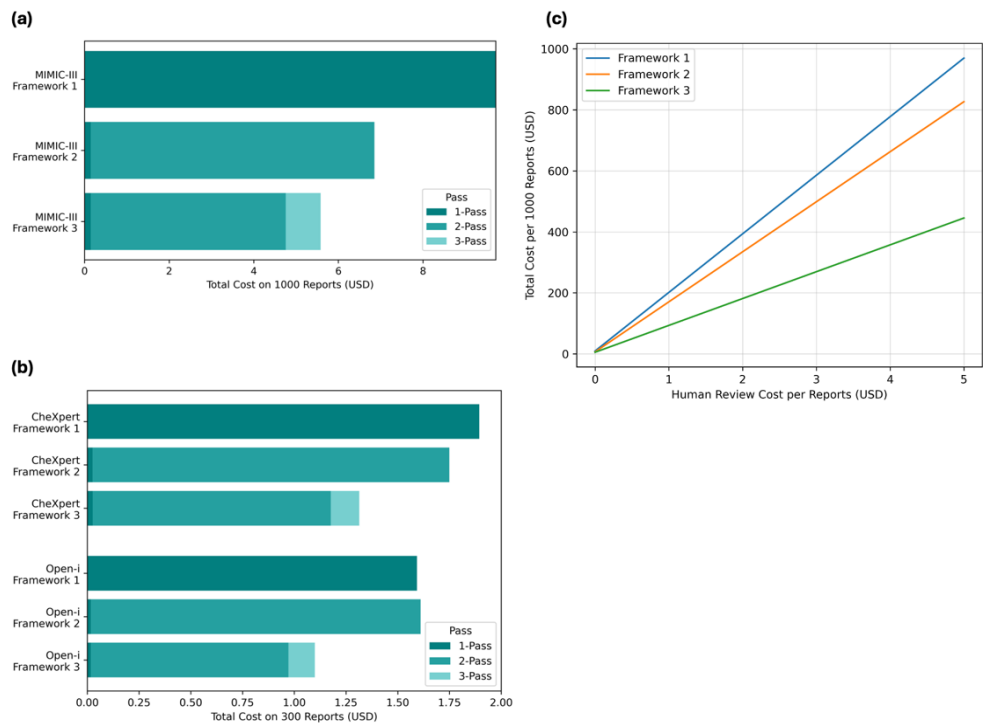

**Figure 5.** Cost analysis of the radiology report error detection frameworks and their component passes. (A) Model-only inference cost per 1,000 reports in the MIMIC-III dataset. (B) Corresponding inference cost for 300 reports in the CheXpert and Open-i datasets. (C) Total cost for processing 1,000 MIMIC-III reports plotted against the human review fee, ranging from 0 to 5 USD per report. Abbreviation: USD = United States dollars.

**Discussion**

The proposed three-pass LLM framework improved precision and reduced inference costs without compromising error detection capability on datasets that approximate real-world error prevalence. On the MIMIC dataset, a PPV of 16% was achieved—more than twice that of a single-prompt, single-extraction baseline—while maintaining the absolute number of true positives. The computational cost decreased by 42.7 % (USD 9.72 vs. USD 5.57 per 1,000 reports), and the number of alerts requiring human review declined by 54.2 % (192 vs. 88). These improvements remained robust across two independent datasets and within modality-specific subgroups.

Because radiology report errors occur infrequently in routine clinical practice, their low prevalence poses inherent challenges for study design and performance evaluation. To address these challenges, previous studies have often relied on synthetic error injection, based on the assumption that prevalence does not influence the sensitivity or specificity of the model(1,10,11). However, this approach has notable limitations. Specifically, synthetic error injection may introduce bias in performance evaluation and error characterization, as the distribution and nature of injected errors may not accurately reflect real-world conditions. Furthermore, artificially inflating error prevalence can substantially overestimate the PPV, thereby misrepresenting the practical utility of the model in real-world scenarios. A low PPV—implying a high rate of false alarms—can increase the workload for radiologists and introduce potential biases for researchers conducting quality assurance on curated datasets. Kim et al. demonstrated that few-shot prompting could improve GPT-4's PPV to 0.12 on a dataset without injected errors; however, this improvement was derived from a post hoc analysis that re-prompted only those cases previously identified as false positives, limiting the generalizability of the findings.

Thus, the proposed multi-pass architecture improves precision in real-world settings. This improvement is driven by two key components. First, a preprocessing LLM transforms raw radiology reports into cleaned, structured JSON before passing them to the primary LLM. During prompt tuning, we frequently observed that artifacts—such as embedded metadata, addenda, and page breaks—were misinterpreted as report content, thereby inflating FP rates. The preprocessor mitigates this issue by removing such noise, which not only reduces the likelihood of false positives but also decreases the input size for downstream tasks. However, when applied to the Open-i dataset—which is already extensively cleaned—these preprocessing prompts yielded minimal benefit. This suggests that preprocessing strategies should be tailored to the reporting conventions and dataset characteristics of each institution to achieve optimal performance.

Second, the framework employs a detector–verifier cascade. When the detector is prone to false positives, separating detection and verification into two distinct steps allows the LLMs to complement each other: the detector prioritizes sensitivity, while the verifier enhances specificity. This arrangement parallels the tiered double-reading workflow commonly used in radiology; however, in this framework, the two LLMs perform the initial "double read," and a human radiologist provides the final adjudication—effectively constituting a tiered triple read (23). Prior evidence supports the benefits of task separation: in one study, two GPT-4 prompts for radiology report error detection were compared, revealing a trade-off between sensitivity and specificity, while the overall F1 score remained constant(3). This suggests that, for error detection—where high sensitivity is essential—a two-stage cascade that first maximizes sensitivity and then applies a highly specific verifier offers a more effective balance between error detection and alert fatigue.

By streamlining the review process, the proposed multi pass framework aims to enhance radiologist–AI collaboration in report quality assurance while concurrently addressing environmental sustainability(24). It reduces alert fatigue, restructures flagged reports into a review-friendly format, and displays the model's error reasoning alongside each corresponding finding—thereby lowering the cognitive load for reviewers within a single-screen interface designed for seamless workflow integration. These same efficiencies align with the recommendations of Doo et al.—including energy-efficient model configurations, radiology-specific greenhouse gas calculators, collaborative efforts to reduce redundant models, and optimized data compression—indicating that this framework could serve as a practical example of implementing an energy-reduction roadmap in radiology AI (5).

This study has some limitations. First, because direct measurements of power consumption and carbon emissions of the closed source model were not feasible, we used token processing charge as a surrogate. This approach was chosen to comparatively evaluate the superiority between frameworks, and actual measurements were beyond the scope of this study. Future studies should aim to validate actual energy usage in real-world deployment scenarios. Second, although the PPV improved to 16%, the framework still generates an excessive number of alerts for a busy clinical workflow. In this study, typographical errors, along with all error candidates that could not be confirmed using the corresponding images were conservatively classified as false positives; therefore, the reported PPV likely represents a lower bound. Even with this conservative estimate, the current precision remains insufficient for fully autonomous AI adoption. Many false positives resulted from the framework interpreting individual words too strictly, indicating a limitation in its ability to interpret clinical context effectively. Future studies should focus on further optimization to improve the model's reasoning capabilities by integrating a more comprehensive understanding of the overall clinical scene context.

In conclusion, the multi-pass LLM pipeline improved the precision and efficiency of radiology-report error detection in real-world settings. The framework facilitates AI–radiologist collaboration and offers a potentially more energy-efficient and scalable approach to AI-assisted quality assurance in routine radiology practice.

**References**

1. Gertz RJ, Dratsch T, Bunck AC, Lennartz S, Iuga AI, Hellmich MG, et al. Potential of GPT-4 for detecting errors in radiology reports: Implications for reporting accuracy. Radiology. 2024 Apr;311(1):e232714.

2. Forman HP. Large language models as an inexpensive and effective extra set of eyes in radiology reporting. Radiology. 2024 Apr;311(1):e240844.

3. Kim S, Kim D, Shin HJ, Lee SH, Kang Y, Jeong S, et al. Large-scale validation of the feasibility of GPT-4 as a proofreading tool for head CT reports. Radiology. 2025 Jan;314(1):e240701.

4. Philpotts LE. Advancing artificial intelligence to meet breast imaging needs. Radiology. 2022 Apr;303(1):78–9.

5. Doo FX, Vosshenrich J, Cook TS, Moy L, Almeida EPRP, Woolen SA, et al. Environmental sustainability and AI in radiology: A double-edged sword. Radiology. 2024 Feb;310(2):e232030.

6. Chen X, Yi H, You M, Liu W, Wang L, Li H, et al. Enhancing diagnostic capability with multi-agents conversational large language models. NPJ Digit Med. 2025 Mar 13;8(1):159.

7. Xie Q, Chen Q, Chen A, Peng C, Hu Y, Lin F, et al. Medical foundation large language models for comprehensive text analysis and beyond. NPJ Digit Med. 2025 Mar 5;8(1):141.

8. Patterson D, Gonzalez J, Le Q, Liang C, Munguia LM, Rothchild D, et al. Carbon emissions and large neural network training [Internet]. arXiv [cs.LG]. 2021. Available from: http://arxiv.org/abs/2104.10350

9. Cemri M, Pan MZ, Yang S, Agrawal LA, Chopra B, Tiwari R, et al. Why Do Multi-Agent LLM Systems Fail? [Internet]. arXiv [cs.AI]. 2025. Available from: http://arxiv.org/abs/2503.13657

10. Salam B, Stüwe C, Nowak S, Sprinkart AM, Theis M, Kravchenko D, et al. Large language models for error detection in radiology reports: a comparative analysis between closed-source and privacy-compliant open-source models. Eur Radiol [Internet]. 2025 Feb 20; Available from: http://dx.doi.org/10.1007/s00330-025-11438-y

11. Sun C, Teichman K, Zhou Y, Critelli B, Nauheim D, Keir G, et al. Generative Large language models trained for detecting errors in radiology reports. Radiology. 2025 May;315(2):e242575.

12. Johnson AEW, Pollard TJ, Shen L, Lehman LWH, Feng M, Ghassemi M, et al. MIMIC-III, a freely accessible critical care database. Sci Data. 2016 May 24;3:160035.

13. Irvin J, Rajpurkar P, Ko M, Yu Y, Ciurea-Ilcus S, Chute C, et al. CheXpert: A Large Chest Radiograph Dataset with Uncertainty Labels and Expert Comparison. AAAI. 2019 Jul 17;33(01):590–7.

14. Demner-Fushman D, Kohli MD, Rosenman MB, Shooshan SE, Rodriguez L, Antani

S, et al. Preparing a collection of radiology examinations for distribution and retrieval. J Am Med Inform Assoc. 2016 Mar;23(2):304–10.

15. OpenAI Platform [Internet]. [cited 2025 Jun 11]. Available from: https://platform.openai.com/docs/guides/structured-outputs/examples?api-mode=responses

16. OpenAI Platform [Internet]. [cited 2025 Jun 11]. Available from: https://platform.openai.com/docs/models

17. Briggs AH, O'Brien BJ. The death of cost-minimization analysis? Health Econ. 2001 Mar;10(2):179–84.

18. Krupp L, Geißler D, Lukowicz P, Karolus J. Towards sustainable web agents: A plea for transparency and dedicated metrics for energy consumption [Internet]. arXiv [cs.AI]. 2025. Available from: http://dx.doi.org/10.48550/ARXIV.2502.17903

19. Strubell E, Ganesh A, McCallum A. Energy and policy considerations for deep learning in NLP [Internet]. arXiv [cs.CL]. 2019. Available from: http://arxiv.org/abs/1906.02243

20. Poddar S, Koley P, Misra J, Podder S, Ganguly N, Ghosh S. Towards sustainable NLP: Insights from benchmarking inference energy in large language models [Internet]. arXiv [cs.CL]. 2025. Available from: http://arxiv.org/abs/2502.05610

21. Clopper CJ, Pearson ES. The use of confidence or fiducial limits illustrated in the case of the binomial. Biometrika. 1934;26(4):404–13.

22. Cameron AC, Gelbach JB, Miller DL. Bootstrap-based improvements for inference with clustered errors. Rev Econ Stat. 2008 Aug;90(3):414–27.

23. Suri A. AI as a second reader can reduce radiologists' workload and increase accuracy in screening mammography. Radiol Artif Intell. 2024 Nov;6(6):e240624.

24. Park SH, Langlotz CP. Crucial role of understanding in human-artificial intelligence interaction for successful clinical adoption. Korean J Radiol. 2025 Apr;26(4):287–90.

**Funding information**

MD-PhD/Medical Scientist Training Program through the Korea Health Industry Development Institute, funded by the Ministry of Health & Welfare, Republic of Korea

Korea Health Technology R&D Project through the Korea Health Industry Development Institute (KHIDI), funded by the Ministry of Health & Welfare, Republic of Korea (grant number : HI22C0452).

**Table 1.** Representative cases from the analysis of false positives across three different frameworks

| Report Excerpt | Framework 1 | Framework 2 | Framework 3 | False positive rationale |
|---|---|---|---|---|
| Header: "Comparison: 10/20." "1. Compared to prior study from October 5th, 20, interval increase in..." | Error | No error | No error | Two legitimate comparison dates were interpreted as contradictory. |
| Header: "s/p MVC, s/p chest removal" "status post MVC and chest tube removal, " | Error | No error | No error | Typographic omission of "tube" was mistaken for a clinical conflict. |
| "...image degradation in the low pelvis because of patient's size, but no masses or fluid collections are seen." "osteolytic and mixed osteosclerotic metastases are seen in the pelvic bones, most prominent at the right iliac..." | Error | Error | No error | Separate reporting of the pelvic cavity and pelvic bone was overlooked, and the statements were therefore flagged as contradictory. |
| Chest section: "The heart, pericardium, and great vessels are normal." Abdomen section: "The IVC is markedly compressed; however, remains patent." | Error | Error | No error | Separate reporting of chest and abdomen was overlooked, and the statements were therefore flagged as contradictory. |
| "The liver demonstrates normal morphology without signal dropout...There are numerous ill-defined lesions within the liver which are hypointense to the liver parenchyma..." | Error | Error | No error | Failure to distinguish overall morphology from focal lesions produced a false positive. |

**Table 2.** Positive predictive value among three error detection frameworks across MIMIC-III, CheXpert, and Open-i datasets

| Dataset | Modality | Framework | TP | FP | PPV (95% CI) | p-value[+] | p-value[++] | p-value[+++] |
|---|---|---|---|---|---|---|---|---|
| MIMIC-III | | | | | | | | |
| | Overall | 1 | 12 | 179 | 0.063 (0.033–0.107) | 0.013 | 0.013 | 0.019 |
| | Overall | 2 | 13 | 151 | 0.079 (0.043–0.132) | <.001 | <.001 | |
| | Overall | 3 | 14 | 74 | 0.159 (0.090–0.252) | <.001 | <.001 | |
| | X-ray | 1 | 2 | 17 | 0.105 (0.013–0.331) | 1.000 | | 0.523 |
| | X-ray | 2 | 2 | 16 | 0.111 (0.014–0.347) | 0.286 | | |
| | X-ray | 3 | 2 | 8 | 0.200 (0.025–0.556) | 0.285 | | |
| | Ultrasound | 1 | 3 | 27 | 0.100 (0.021–0.265) | 0.420 | | 0.268 |
| | Ultrasound | 2 | 3 | 23 | 0.115 (0.024–0.302) | 0.040 | | |
| | Ultrasound | 3 | 4 | 14 | 0.222 (0.064–0.476) | 0.034 | | |
| | CT | 1 | 4 | 85 | 0.045 (0.012–0.111) | 0.018 | | 0.089 |
| | CT | 2 | 5 | 64 | 0.072 (0.024–0.161) | 0.013 | | |
| | CT | 3 | 5 | 30 | 0.143 (0.048–0.303) | 0.013 | | |
| | MR | 1 | 3 | 50 | 0.057 (0.012–0.157) | 0.890 | | 0.385 |
| | MR | 2 | 3 | 48 | 0.059 (0.012–0.162) | 0.090 | | |
| | MR | 3 | 3 | 22 | 0.120 (0.025–0.312) | 0.096 | | |
| CheXpert | | | | | | | | |
| | | 1 | 2 | 25 | 0.074 (0.009–0.243) | 0.459 | 0.790 | 0.548 |
| | | 2 | 2 | 19 | 0.095 (0.012–0.304) | 0.390 | 0.790 | |
| | | 3 | 2 | 13 | 0.133 (0.017–0.405) | 0.263 | 0.790 | |
| Open-i | | | | | | | | |
| | | 1 | 2 | 22 | 0.083 (0.010–0.270) | 0.268 | 0.803 | 0.835 |
| | | 2 | 2 | 41 | 0.047 (0.006–0.158) | 0.247 | 0.803 | |
| | | 3 | 2 | 17 | 0.105 (0.013–0.331) | 0.566 | 0.803 | |

[+] Two-sided paired-cluster bootstrap (1000 replicates) p-value — this row compares current framework with the next (row 1: Framework 1 vs Framework 2, row 2: Framework 2 vs Framework 3, row 3: Framework 1 vs Framework 3). [++] Holm-adjusted p-value — same comparisons as above. [+++] Cochran–Armitage trend test p-value. Abbreviations: TP, true positives; FP, false positives; PPV, positive predictive value; CI, confidence interval;

**Table 3.** Absolute true positive rate among three error detection frameworks across MIMIC-III, CheXpert, and Open-i datasets

| Dataset | Modality | Framework | aTPR (95% CI) | p-value[+] | p-value[++] | p-value[+++] |
|---|---|---|---|---|---|---|
| MIMIC-III | | | | | | |
| | Overall | 1 | 0.012 (0.006–0.021) | 1.000 | 1.000 | 0.926 |
| | Overall | 2 | 0.013 (0.007–0.022) | 1.000 | 1.000 | |
| | Overall | 3 | 0.014 (0.008–0.023) | 0.845 | 1.000 | |
| | X-ray | 1 | 0.008 (0.001–0.029) | 1.000 | | 1.000 |
| | X-ray | 2 | 0.008 (0.001–0.029) | 1.000 | | |
| | X-ray | 3 | 0.008 (0.001–0.029) | 1.000 | | |
| | Ultrasound | 1 | 0.012 (0.002–0.035) | 1.000 | | 0.905 |
| | Ultrasound | 2 | 0.012 (0.002–0.035) | 1.000 | | |
| | Ultrasound | 3 | 0.016 (0.004–0.040) | 1.000 | | |
| | CT | 1 | 0.016 (0.004–0.040) | 1.000 | | 0.931 |
| | CT | 2 | 0.020 (0.007–0.046) | 1.000 | | |
| | CT | 3 | 0.020 (0.007–0.046) | 1.000 | | |
| | MR | 1 | 0.012 (0.002–0.035) | 1.000 | | 1.000 |
| | MR | 2 | 0.012 (0.002–0.035) | 1.000 | | |
| | MR | 3 | 0.012 (0.002–0.035) | 1.000 | | |
| CheXpert | | | | | | |
| | | 1 | 0.007 (0.001–0.024) | 1.000 | 1.000 | 1.000 |
| | | 2 | 0.007 (0.001–0.024) | 1.000 | 1.000 | |
| | | 3 | 0.007 (0.001–0.024) | 1.000 | 1.000 | |
| Open-i | | | | | | |
| | | 1 | 0.007 (0.001–0.024) | 1.000 | 1.000 | 1.000 |
| | | 2 | 0.007 (0.001–0.024) | 1.000 | 1.000 | |
| | | 3 | 0.007 (0.001–0.024) | 1.000 | 1.000 | |

[+] McNemar's test p-value — this row compares current framework with the next (row 1: Framework 1 vs Framework 2; row 2: Framework 2 vs Framework 3; row 3: Framework 1 vs Framework 3). [++] Holm-adjusted p-value — same comparisons as above. [+++] Cochran's Q test p-value. Abbreviations: aTPR, absolute true positive rate; CI, confidence interval

**Table 4.** Inference cost per pass among three error detection frameworks across MIMIC-III, CheXpert, and Open-i datasets

| Dataset | Framework | # Passes | Model | N calls | Input tokens (×10³) | Output tokens (×10³) | Input/call | Output/call | Cost (USD) |
|---|---|---|---|---|---|---|---|---|---|
| MIMIC-III | | | | | | | | | |
| | 1 | 1 | o3 | 1000 | 868.532 | 25.925 | 868.53 | 25.92 | 9.7223 |
| | 2 | 1 | 4.1-nano | 1000 | 681.532 | 187.901 | 681.53 | 187.9 | 0.1433 |
| | 2 | 2 | o3 | 1000 | 575.901 | 23.72 | 575.9 | 23.72 | 6.7078 |
| | 3 | 1 | 4.1-nano | 1000 | 681.532 | 187.071 | 681.53 | 187.07 | 0.143 |
| | 3 | 2 | o3 | 1000 | 369.071 | 23.089 | 369.07 | 23.09 | 4.6143 |
| | 3 | 3 | o3 | 88 | 50.175 | 8.023 | 570.17 | 91.17 | 0.8227 |
| CheXpert | | | | | | | | | |
| | 1 | 1 | o3 | 300 | 166.082 | 5.837 | 553.61 | 19.46 | 1.8943 |
| | 2 | 1 | 4.1-nano | 300 | 109.982 | 35.635 | 366.61 | 118.78 | 0.0253 |
| | 2 | 2 | o3 | 300 | 152.035 | 5.087 | 506.78 | 16.96 | 1.7238 |
| | 3 | 1 | 4.1-nano | 300 | 109.982 | 36.58 | 366.61 | 121.93 | 0.0256 |
| | 3 | 2 | o3 | 300 | 91.18 | 5.968 | 303.93 | 19.89 | 1.1505 |
| | 3 | 3 | o3 | 15 | 7.579 | 1.539 | 505.27 | 102.6 | 0.1374 |
| Open-i | | | | | | | | | |
| | 1 | 1 | o3 | 300 | 139.682 | 4.904 | 465.61 | 16.35 | 1.593 |
| | 2 | 1 | 4.1-nano | 300 | 83.582 | 20.498 | 278.61 | 68.33 | 0.0166 |
| | 2 | 2 | o3 | 300 | 136.898 | 5.615 | 456.33 | 18.72 | 1.5936 |
| | 3 | 1 | 4.1-nano | 300 | 83.582 | 20.546 | 278.61 | 68.49 | 0.0166 |
| | 3 | 2 | o3 | 300 | 75.146 | 5.091 | 250.49 | 16.97 | 0.9551 |
| | 3 | 3 | o3 | 19 | 6.719 | 1.497 | 353.63 | 78.79 | 0.1271 |

# Appendix: A Multi-Pass Large Language Model Framework for Precise and Efficient Radiology Report Error Detection

## Supplementary Materials and Methods

**Prompt and parameter**

**1) 1st Pass LLM**

***Prompt:***

**Tasks**

1. Extract only content belonging to the *Findings* section (detailed observations) and the *Impression* / *Conclusion* / *Opinion* section (diagnostic interpretation).
2. If there is an *Addendum* or *Correction* section:
    - Sentences that amend Findings ⇒ append to Findings.
    - Sentences that amend the diagnostic interpretation ⇒ append to Impression.
    - If ambiguous, append to Impression.
    - Also, clearly mark it as "Addendum".
3. Discard every other section (history, technique, timestamps, signatures, headers, billing codes, etc.).
4. Replace every explicit calendar date with the literal token **[DATE]**.
5. Replace PHI with the literal token **[PHI]**.

(Output must follow the JSON schema exactly.)

{"Findings":"~", "Impression":"~"}

If either the Findings or Impression section is missing, set the corresponding value to "N/A".

***Parameter:***

*Json schema*:

```
{
    "name": "preprocessing",
    "strict": True,
    "schema": {
        "type": "object",
        "properties": {
            "findings":    {"type": "string"},
            "impression": {"type": "string"},
        },
        "required": ["findings", "impression"],
        "additionalProperties": False,
    },
```

}

Other parameters are set to defaults.

## 2) 2nd Pass LLM

### Prompt:

**Tasks**

Identify clinically significant errors based on the content within the provided radiology report.

1. Please read through the entire radiology report and understand the clinical scenario.
2. Identify any clinically significant errors in the report.
3. Limit errors to parts identifiable without images:
- Internal factual inconsistencies: Directly conflicting statements within the same radiology reports (e.g., conflicting laterality, measurements).
- Objective misinterpretations: Interpretations clearly and objectively contradicted by explicit statements in the Findings and Impression sections of the same radiology report.

(Output must follow the JSON schema exactly.)
If no error is found, return the JSON with
"error": "no error", "error_reason": "N/A".
If an error is found, return the JSON with
"error": "(cite erroneous statement from the report)", "error_reason": "(concise explanation; utilize quotes if necessary)".

### Parameter:

Json schema:
{
    "name": "error_report",
    "strict": True,
    "schema": {
        "type": "object",
        "properties": {
            "error": {"type": "string"},
            "error_reason": {"type": "string"},
        },
        "required": ["error", "error_reason"],
        "additionalProperties": False,
    },
}

Other parameters are set to defaults.

## 3) 3rd Pass LLM

*Prompt:*

You will receive: 'radiology report JSON' and 'candidate error JSON'.

**Tasks**
Decide whether `candidate error JSON` is a TRUE clinically-significant internal error within `radiology report JSON` or a FALSE POSITIVE.

Guidelines to confirm an error:
- Objectivity: A true ERROR must be objectively incorrect, factually contradictory, or undeniably inaccurate.
- Clarity: The error must be so clear and obvious that ALL trained radiologists or medical professionals would unanimously agree it is incorrect.
- Clinical importance, differences in judgment, or disagreements about what should be included in the Impression or Findings DO NOT qualify as errors.
- Only contradictions or inaccuracies explicitly within the radiology report itself can qualify as errors. Differences between the report content and
    provided clinical information, patient history, or external context must NEVER be considered errors.

(Output must follow the JSON schema exactly.)
If determined to be a FALSE POSITIVE, return JSON with:
"error": "no error", "error_reason": "N/A".

*Parameter:*

*Json schema*:
```
{
    "name": "error_report",
    "strict": True,
    "schema": {
        "type": "object",
        "properties": {
            "error": {"type": "string"},
            "error_reason": {"type": "string"},
        },
        "required": ["error", "error_reason"],
        "additionalProperties": False,
    },
}
```

Other parameters are set to defaults.

# Efficiency evaluation

## 1) Notation and units

We let $N_{\text{in},k}$ and $N_{\text{out},k}$ denote the numbers of input and output tokens, respectively, for the *k-th* model pass. The constants $P_{\text{in}}$ and $P_{\text{out}}$ represent the vendor's charge per input and output token. The set $E$ collects every report that was flagged for human review—regardless of whether the flag was ultimately adjudicated as a true or a false error—and $|E|$ therefore equals the total count of such flags. Finally, $C_{\text{review}}$ denotes the mean fee paid to a radiologist for adjudicating a single flagged report. A complete list of notation is provided in Supplementary Table 1, and the model-specific prices are summarised in Supplementary Table 2.

**Supplementary Table 1.** Definitions of symbols and corresponding units used in the cost-efficiency analysis.

| Symbol | Definition | Unit |
|---|---|---|
| $N_{\text{in},k}$ | Input tokens processed in model pass $k$ | tokens |
| $N_{\text{out},k}$ | Output tokens generated in model pass $k$ | tokens |
| $P_{\text{in}}$ | Price per input token (model-specific) | USD / token |
| $P_{\text{out}}$ | Price per output token (model-specific) | USD / token |
| $E$ | Set of all reports that the model flagged for human review | |
| $C_{\text{review}}$ | Average fee paid for reviewing one flagged report | USD / report |

**Supplementary Table 2.** Supplementary Table 2. Processing prices of OpenAI large-language models (as of 10 May 2025)

| Model | Input price (USD / 1 K tokens) | Output price (USD / 1 K tokens) |
|---|---|---|
| **gpt-4.1-nano** | $0.10 | $0.40 |
| **o3** | $10.00 | $40.00 |
| **o4-mini** | $1.10 | $4.40 |

## 2) Derivation of cost components

The **model-related cost** accumulates over all $K$ passes:

$$C_{\text{model}} = \sum_{k=1}^{K} \left( N_{\text{in},k} P_{\text{in}} + N_{\text{out},k} P_{\text{out}} \right) \quad - \text{Eq. S1}$$

The **human-review cost** converts is calculated by multiplying the number of reports flagged for review by the average fee paid per reviewed report.

$$C_{\text{human}} = |E| \, C_{\text{review}} \quad - \text{Eq. S2}$$

The **total expenditure** is the sum of two components:

$$C_{\text{total}} = C_{\text{model}} + C_{\text{human}} \quad - \text{Eq. S3}$$

# Supplementary Results

**Supplementary Table 3.** True-positive detection errors identified by Framework 3 in the MIMIC-III, CheXpert, and Open-i datasets

| Report | Description |
|---|---|
| **MIMIC-III** | |
| FINDINGS: AP single view of the chest has been obtained with patient in sitting semi-upright position.<br>...<br>A right-sided PICC line is now identified, seen to terminate overlying the right-sided mediastinal structures at the level 3 cm below the carina. This is compatible with the lower third of the _IVC_.<br>... | CHEST PORT. LINE PLACEMENT |
| ...<br>The focal ill-defined opacity in this region noted on the prior film of [**2187-8-10**], has resolved. A jugular CV line is in proximal SVC. No pneumothorax. _Bilateral pleural effusions._<br><br>IMPRESSION: Resolution of peripheral ill-defined opacity in left upper lobe. Persistent spiculated density in this location. _No pneumothorax or bilateral pleural effusions._ ?? Scarring ?? right lung apex. | CHEST (PORTABLE AP) |
| There is an _echogenic vessel in the right thalamus_. This is seen on the sagittal views. It is not changed from the prior examination and likely represents a calcified lenticulostriate vessel.<br>...<br>IMPRESSION: Calcified v_essel in the left thalamus_ is unchanged. Examination is normal otherwise. | NEONATAL HEAD PORTABLE |
| ...<br>On the _right, moderate plaque is seen at the upper portion of the common carotid artery (CCA) as well as at the origins of the internal (ICA) and external (ECA) carotid arteries. The peak systolic velocities (PSVs) are 160, 100 and 148 cm/sec, respectively._ The _right ICA to CCA PSV ratio is 1.6._<br>...<br>CONCLUSION:<br>1. Bilateral moderate plaque in the ICAs, with associated luminal narrowing of 60 to 69% (in diameter) on the right and 40 to 59% on the left.<br>... | CAROTID SERIES COMPLETE |
| ...<br>On the left, there is wall thickening of the common carotid artery, with some echogenic deposits with posterior acoustic shadowing (calcifications). The stent is visualized. T_he peak systolic velocities in the internal, external common carotid arteries are 84, 83 and 269 cm/second, respectively. The left internal to common carotid artery velocity ratio is 1.01._<br><br>...<br><br>CONCLUSION: No significant change as compared to the two previous examinations. | CAROTID SERIES COMPLETE |
| ...<br>   On the right systolic/end diastolic velocities of the ICA proximal, mid and distal respectively are 87/15, 88/22, 72/17, cm/sec. CCA peak systolic velocity is 110/14 cm/sec. ECA peak systolic velocity is 89 cm/sec. The ICA/CCA ratio is .80. *These findings are consistent with <u>no stenosis.</u>*<br>...<br><br>   Impression: <u>*Right ICA stenosis.*</u><br>   Left ICA stenosis 40-59%. | CAROTID SERIES COMPLETE |
| ...<br>   No newly developed mass effect with persistent <u>*approximately 9 mm leftward midline deviation from known extensive right temporal mass.*</u> No other areas of high attenuation that would be concerning for hemorrhage. There is also noted to be sulcal effacement by this mass with associated and compression of the right lateral ventricle.<br>...<br><br>   IMPRESSION: Several areas of high attenuation seen within the right temporal partial lobectomy bed done could be related to mild post op hemorrhage . . No new mass effect with <u>*stable rightward midline shift by approximately*</u> stable leftward midline shift by approximately 9 mm.<br>... | CT HEAD W/O CONTRAST |
| ...<br>   There is periventricular hypoattenuation consistent with small vessel ischemic disease. The right mastoid, middle ear and <u>*right sphenoid are opacified consistent with otitis media as well as mastoiditis.*</u> Also noted is mild opacification of the left mastoid air cells as well as the right sphenoid. This is likely secondary to endotracheal intubation.<br>...<br><br>   IMPRESSION:<br>...<br>   2. Opacification of the right mastoid, <u>*both sphenoid sinuses*</u> and right middle ear cavity likely representing otitis media and mastoiditis secondary to ETT placement. | CT HEAD W/O CONTRAST |
| ...<br>   Please note that this exam was not tailored for subdiaphragmatic evaluation. Limited evaluation of the included upper abdomen displayed some radiopaque substances within the gastric lumen likely related to pill ingestion and a <u>*large conglomerate 26 x 51-mm calcified mass within the peritoneum*</u> displacing the adjacent diaphragmatic crus which is most consistent with calcified nodes from treated lymphoma.<br>...<br><br>   IMPRESSION:<br>...<br>   2. Mild post-radiation changes involving the paramediastinal upper lobes consistent with known treated lymphoma. Large conglomerate calcified <u>*retroperitoneal mass*</u> also resultant of treated lymphoma.<br>... | CT CHEST W/O CONTRAST |
| ...<br>There is a new 1.2 x 0.9 cm _nodule in the superior segment of the left lower lobe_ (2:21).There is also a new area of ground-glass opacity in the medial aspect of the right middle lobe measuring approximately 1.0 x 0.3 cm (3:28). Previously identified pulmonary nodules in bilateral lungs are otherwise stable (3:24, ,30, 34, 35, 46, 50) .<br>...<br><br>IMPRESSION:<br>1. New 1.2 x 0.9 cm nodule in the _superior segment of the left upper lobe_ and new area of ground-glass opacity in the medial aspect of the right middle lobe, as well as multiple stable pulmonary nodules likely metastatic in nature.<br>... | CT CHEST W/O CONTRAST |
| ...<br>Interval evolution of bilateral cerebellar hemispheric infarcts. Unchanged small right frontal intraparenchymal hemorrhage surrounding a right frontal approach ventriculostomy catheter which ends at the foramen of [**Last Name (un) 7030**].   The known diffuse bilateral subarachnoid hemorrhages as well as the     intraventricular hemorrhage layering in the occipital horns have slightly decreased in density. _New acute intracranial hemorrhage and no acute cerebral infarction._   Unchanged mucosal thickening of the left maxillary sinus, sphenoid sinus,   ethmoid and frontal sinuses. The mastoid air cells are clear.<br><br>IMPRESSION:<br>...<br>5. _No new acute intracranial hemorrhage or new acute infarction._ | CT HEAD W/O CONTRAST |
| MR brain with contrast: Within the [**Doctor Last Name 37**] matter of the _left parietal lobe is a  small thick rim enhancing 7 mm round lesion_ with moderate amount of associated    vasogenic edema. Signal is slightly hyperdense to [**Doctor Last Name 37**] matter on T2 imaging    with a hypodense surrounding rim.<br>...<br>IMPRESSION:<br>1. Thick rim enhancing _7 mm right parietal lesion_ with associated edema that represents abscess versus neoplasm. Correlation with outside CT to determine presence of calcification is advised, and if access to outside CT is not available, reimaging is advised for further characterization.<br>... | MR HEAD W & W/O CONTRAST |
| ...<br>The right Posterior Inferior Cerebellar artery is visualized on the prior CTA; left Anterior inferior cerebellar artery is not seen. The P1 segments of the _posterior cerebellar arteries_ are hypolastic with prominent posterior communicating arteries on both sides representing fetal pattern.<br><br>IMPRESSION:<br>1. Multiple small acute infarcts within the left MCA territory with associated stenosis of the M2 branches of the left middle cerebral artery, better evaluated on prior CTA.<br>2. Mild thickening of the ligaments posterior to the dens without cord compression- can be degenerative or inflammatory- to correlate with past history. | MR HEAD W/O CONTRAST |

| | MR HEAD W/O CONTRAST |
|---|---|
| ...<br>   There is mildly increased signal, in the cortex in the left frontal lobe, (series 6, image 20), superiorly near the vertex, with increased signal on the diffusion-weighted sequence. In addition, *there are two punctate foci, noted adjacent to the cortex, in the left frontal and in the right parietal lobes* (series 302, image 22). These do not have definitively convincing decreased signal on the ADC sequence.<br>...<br><br>   IMPRESSION:<br>   1. Small areas of increased signal on the FLAIR and the DWI in the left frontal lobe, the *punctate focus in the left parietal lobe*, which may relate to acute-subacute infarction or edema. Given their small size, accurate assessment is somewhat limited, as these are not clearly identifiable on the ADC sequence. To correlate clinically and if necessary, followup can be considered.<br>... | |
| **CheXpert** | |
| ...<br>   Right internal jugular sheath with central line, endotracheal tube, nasogastric tube, and feeding tube are unchanged. There is no evidence of pneumothorax. *Compared with the prior examination, there has been slight increase in interstitial pulmonary edema*. However, there is decreased aeration at the left base.<br>...<br><br>   IMPRESSION:<br>   ...<br>   3. *DECREASED INTERSTITIAL PULMONARY EDEMA.* | |
| ...<br>The lung parenchyma is clear. There *are no pleural or significant bony abnormalities.*<br><br>   IMPRESSION:<br>   ...<br>    *Bilateral moderate pleural effusions.*<br>     No focal infiltrate. | |
| **Open-i** | |
| ...<br>There has been interval sternotomy with intact midline sternotomy XXXX. *The heart is near top normal in size* with unfolding of the aorta.<br>...<br>   IMPRESSION:<br>   *Cardiomegaly*, however no acute cardiopulmonary findings. | |
| ...<br>   *The cavity and the left upper lobe has decreased in size.*<br>   Bilateral apical bullae and parenchymal scars are unchanged.<br>   ...<br><br>   IMPRESSION:<br>   Bullous disease and upper lobe scars. *Decreasing right upper lobe cavity.* | |

**Supplementary Table 4.** Positive predictive values between o3 and o4-mini models in the MIMIC-III dataset

| Dataset | Modality | Framework 3 o3 PPV (95% CI) | Framework 3 o4-mini PPV (95% CI) | p-value[+] |
|---|---|---|---|---|
| MIMIC-III | | | | |
| | Overall | 0.159 (0.097-0.250) | 0.081 (0.047-0.136) | <.001 |
| | X-ray | 0.200 (0.057-0.510) | 0.091 (0.025-0.278) | 0.278 |
| | Ultrasound | 0.222 (0.090-0.452) | 0.160 (0.064-0.347) | 0.097 |
| | CT | 0.143 (0.063-0.294) | 0.075 (0.030-0.179) | 0.015 |
| | MR | 0.120 (0.042-0.300) | 0.042 (0.012-0.140) | 0.094 |

[+] Two-sided paired cluster bootsrap (1000 replicates) p-value. Abbreviations: TP, true positives; FP, false positives; PPV, positive predictive value; CI, confidence interval;

**Supplementary Table 5.** Absolute true-positive rates between o3 and o4-mini models in the MIMIC-III dataset

| Dataset | Modality | LLM3_o3 aTPR (95% CI) | LLM3_o4-mini aTPR (95% CI) | p-value[+] |
|---|---|---|---|---|
| MIMIC-III | | | | |
| | Overall | 0.014 (0.008-0.023) | 0.012 (0.007-0.021) | 0.693 |
| | X-ray | 0.008 (0.002-0.029) | 0.008 (0.002-0.029) | 1.000 |
| | Ultrasound | 0.016 (0.006-0.040) | 0.016 (0.006-0.040) | 1.000 |
| | CT | 0.020 (0.009-0.046) | 0.016 (0.006-0.040) | 0.737 |
| | MR | 0.012 (0.004-0.035) | 0.008 (0.002-0.029) | 0.653 |

[+] McNemar test p-value. Abbreviations: aTPR, absolute true-positive rate; CI, confidence interval;